\documentclass[conference]{IEEEtran}
\usepackage[pdftex]{graphicx}
\usepackage{cite}
\usepackage{amsmath, amssymb}
\usepackage{subfigure}
\usepackage[font = footnotesize]{caption}
\usepackage{multirow}

\renewcommand\IEEEkeywordsname{Keywords}

\begin{document}
\title{Fast Bayesian Restoration of Poisson Corrupted Images with INLA}

\author{
    \IEEEauthorblockN{
        Takahiro Kawashima\IEEEauthorrefmark{1},
        Hayaru Shouno\IEEEauthorrefmark{1}
    }
    \IEEEauthorblockA{
        \IEEEauthorrefmark{1}
        Graduate School of Informatics and Engineering,
        The University of Electro-Communications\\
        1-5-1 Chofugaoka, Chofu, Tokyo, 182-8585, Japan\\
        Email: kawashima@uec.ac.jp, shouno@uec.ac.jp
    }
}

\maketitle

\begin{abstract}
    Photon-limited images are often seen in fields such as medical imaging.
    Although the number of collected photons on an image sensor statistically follows Poisson distribution,
    this type of noise is intractable, unlike Gaussian noise.
    In this study, we propose a Bayesian restoration method of Poisson corrupted image
    using Integrated Nested Laplace Approximation (INLA),
    which is a computational method to evaluate marginalized posterior distributions of latent Gaussian models (LGMs).
    When the original image can be regarded as ICAR (intrinsic conditional auto-regressive) model reasonably,
    our method performs very faster than well-known ones such as loopy belief propagation-based method and
    Markov chain Monte Carlo (MCMC) without decreasing the accuracy.
\end{abstract}
\ \\
\begin{IEEEkeywords}
    Probabilistic Image Processing, Image Restoration, Bayesian Modeling, Integrated Nested Laplace Approximation
\end{IEEEkeywords}

\section{Introduction}
Estimating the original image from a noisy observation is
one of the representative issues in the field of image processing.
So far, a lot of image denoising methods have been proposed, but many of them
such as Wiener filter require the assumption of additive white Gaussian noise.
However, for example, in medical X-ray imaging systems,
the number of detected photons stochastically fluctuates following Poisson distribution\cite{boyat_review_2015}.
\par
From a Bayesian viewpoint, it is natural to adopt Poisson observation modeling explicitly,
and then solve the inverse problem to deal with such images.
In fact, some Bayesian methods to restore Poisson corrupted images have been already proposed.
Le et al. introduced a variational model with total-variation regularization\cite{le_variational_2007}.
Lefkimmiatis et al. applied quadtree decomposition and then estimated parameters using
expectation-maximization (EM) algorithm to handle Poisson noise\cite{lefkimmiatis_bayesian_2009}.
Shouno derived the fixed point equations of loopy belief propagation (LBP) by
approximating Poisson distribution with binomial distribution\cite{shouno_bayesian_2015}.
Furthermore, Tachella et al. compared the performances of multiple MCMC methods
to restore Poisson corrupted images \cite{tachella_bayesian_2018}.
\par
These methods are frequently used to evaluate the posterior (or its point estimate) of each pixel.
In more general applications, Markov chain Monte Carlo (MCMC) methods succeeded due to their
usability and accuracy.
On the other hand, recently, integrated nested laplace approximation (INLA) is proposed by Rue et al.
\cite{rue_approximate_2009},
then the validity has been reported mainly in the fields of spatial statistics and epidemiology
\cite{rue_bayesian_2017}.
For example, INLA is used disease mapping\cite{schrodle_spatio_2011},
analysing spatio-temporal evolutions of Ebola virus\cite{santermans_spatiotemporal_2016},
and Bayesian outbreak detection\cite{salmon_bayesian_2015}.
Also there is a similar research to ours, INLA-based super-resolution method has been proposed
\cite{camponez_super-resolution_2012}.
\par
INLA is a powerful framework to evaluate posterior marginals on latent Gaussian models (LGMs).
LGM is a class of statistical models which includes a lot of standards,
for example, auto-regressive (AR), conditional auto-regressive (CAR),
and generalized linear mixture (GLM) models\cite{rue_bayesian_2017, opitz_latent_2017}.
INLA consists of two layers of units: latent variables following Gaussian Markov random field (GMRF) and
their observations.
Latent variables requires normality on their interactions, but observation processes do not.
INLA is based on Laplace approximations and numerical integrations as the name suggests.
As the interface for using INLA, \texttt{R-INLA} is provided by \texttt{R-INLA} project
\footnote{http://www.r-inla.org/}.
\par
In this study, we try applying INLA to image restoring as the application to the new field,
through dealing with Poisson corrupted images.
When the original image seems to be reasonable to assume intrinsic CAR (ICAR) model,
which means almost of adjacent pixels do not vary sharply,
the proposed method obtained equal to the result by an MCMC simulation rapidly.
\par
This paper organized as following: first, we give a brief description of INLA in Section 2.
Then, we explain our model to infer the original images in Section 3,
and to evaluate proposed method we show configurations and results of
the computational simulations in Section 4.
At last, in Section 5, we summarize and conclude about our method.

\section{Methodology}
\subsection{Latent Gaussian Models}
A latent Gaussian Models (LGM) is a class of statistical models,
which include many commonly used statistical models
such as auto-regressive (AR) model, conditional auto-regressive (CAR) model,
and generalized linear mixture (GLM) model.\par
\begin{figure}[t]
    \centering
    \includegraphics[width = 0.8\columnwidth]{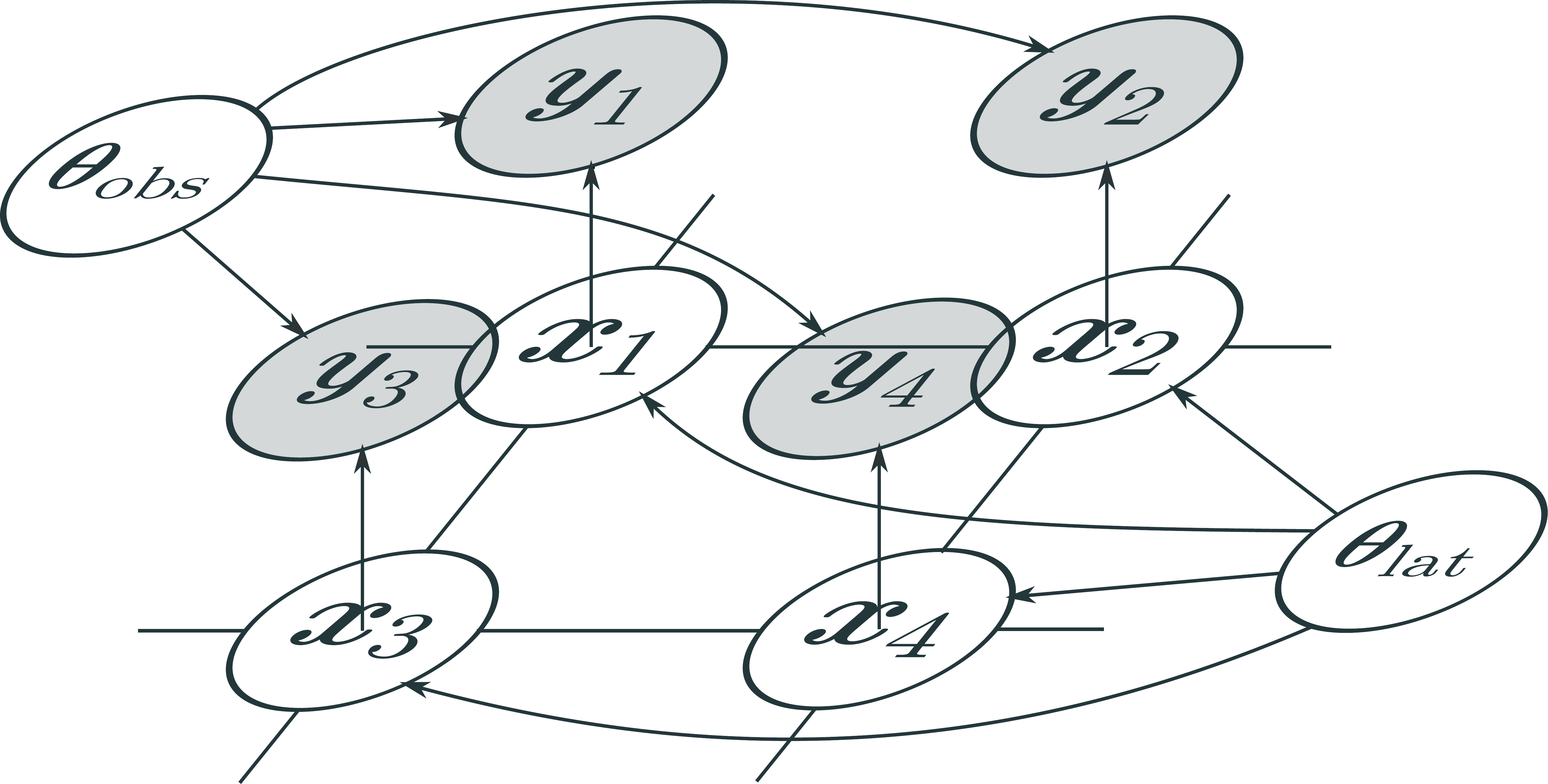}
    \caption{An illustrated outline of a LGM. Highlighted nodes denote observated variables.
    Latent variables compose GMRF which is parameterized by $\boldsymbol{\theta}_{\mathrm{lat}}$ and
    observations should be independent when latent variables $\boldsymbol{x}$ and
    hyperparameters $\boldsymbol{\theta}_{\mathrm{obs}}$
    are given.}
    \label{fig:lgm}
\end{figure}

Fig.\ref{fig:lgm} shows the outline of an LGM.
In LGMs, latent variables
$\boldsymbol{x} = (\boldsymbol{x}_1, \boldsymbol{x}_2, \dots, \boldsymbol{x}_n)^{\top}$
follow Gaussian Markov random field (GMRF)
\begin{IEEEeqnarray}{c}
    p(\boldsymbol{x}) =
    \mathcal{N}(\boldsymbol{x} | \boldsymbol{\mu}(\boldsymbol{\theta}_{\mathrm{lat}}),
    \boldsymbol{\Sigma}^{-1}(\boldsymbol{\theta}_{\mathrm{lat}})),
\end{IEEEeqnarray}
where $\boldsymbol{\theta}_{\mathrm{lat}}$ is a set of hyperparameters of latent GMRF.
In addition, mean vector $\boldsymbol{\mu}$ and precision matrix $\boldsymbol{\Sigma}^{-1}$
are parameterized by $\boldsymbol{\theta}_{\mathrm{lat}}$.\par
Observations
$\boldsymbol{y} = (\boldsymbol{y}_1, \boldsymbol{y}_2, \dots, \boldsymbol{y}_n)^{\top}$
should be conditionally independent and identically distributed
on a parameter set $\boldsymbol{\theta}_{\mathrm{obs}}$
\begin{IEEEeqnarray}{c}
    p(\boldsymbol{y}) =
    \prod^{n}_{i = 1} p(\boldsymbol{y}_i | \boldsymbol{x}, \boldsymbol{\theta}_{\mathrm{obs}}),
\end{IEEEeqnarray}
and conditional distribution
$p(\boldsymbol{y}_i | \boldsymbol{x}, \boldsymbol{\theta}_{\mathrm{obs}})$
must be belong to exponential family.
Hereafter, for simplicity,
we will put hyperparameter sets into $\boldsymbol{\theta}$,
\begin{IEEEeqnarray}{c}
    \boldsymbol{\theta} = \{\boldsymbol{\theta}_{\mathrm{lat}}, \boldsymbol{\theta}_{\mathrm{obs}}\}.
\end{IEEEeqnarray}

\subsection{Integrated Nested Laplace Approximation}
Integrated Nested Laplace Approximation (INLA) is a fast and accurate
method to approximate posterior distributions of LGMs\cite{rue_approximate_2009}.
Especially when the number of hyperparamters is small
(empirically $|\boldsymbol{\theta}| \lesssim 5$\cite{rue_bayesian_2017}),
INLA performs very faster than Markov chain Monte Carlo (MCMC).\par
In LGMs, our aim is to get marginalized posteriors of hyperparameters
\begin{IEEEeqnarray}{c}
    p(\boldsymbol{\theta} | \boldsymbol{y})
    \propto \cfrac{p(\boldsymbol{\theta}) p(\boldsymbol{x} | \boldsymbol{\theta})
                   p(\boldsymbol{y} | \boldsymbol{x}, \boldsymbol{\theta})}
    {p(\boldsymbol{x} | \boldsymbol{\theta}, \boldsymbol{y})}
    \label{posterior_hyper}
\end{IEEEeqnarray}
and latent variables
\begin{IEEEeqnarray}{c}
    p(\boldsymbol{x}_i | \boldsymbol{y})
    = \int p(\boldsymbol{x}_i | \boldsymbol{y}, \boldsymbol{\theta})
    p(\boldsymbol{\theta} | \boldsymbol{y}) \mathrm{d\boldsymbol{\theta}}.
    \label{posterior_latent}
\end{IEEEeqnarray}

In equation \eqref{posterior_hyper}, each term in the numerator can be calculated
by forward computation, but the denominator can not.
Hence, first we apply Laplace (or simplified Laplace,
which can fit with a skew normal distribution\cite{azzalini_statistical_1999, rue_approximate_2009})
approximation to the denominator
\begin{IEEEeqnarray}{c}
    \tilde{p}(\boldsymbol{\theta} | \boldsymbol{y})
    \propto \cfrac{p(\boldsymbol{\theta}) p(\boldsymbol{x} | \boldsymbol{\theta})
                   p(\boldsymbol{y} | \boldsymbol{x}, \boldsymbol{\theta})}
                  {\tilde{p}(\boldsymbol{x} | \boldsymbol{\theta}, \boldsymbol{y})},
\end{IEEEeqnarray}
where $\tilde{p}(\cdot)$ denotes an approximation of $p(\cdot)$.\par
\begin{figure}[t]
    \centering
    \subfigure[grid]{
        \includegraphics[width = 0.46\columnwidth]{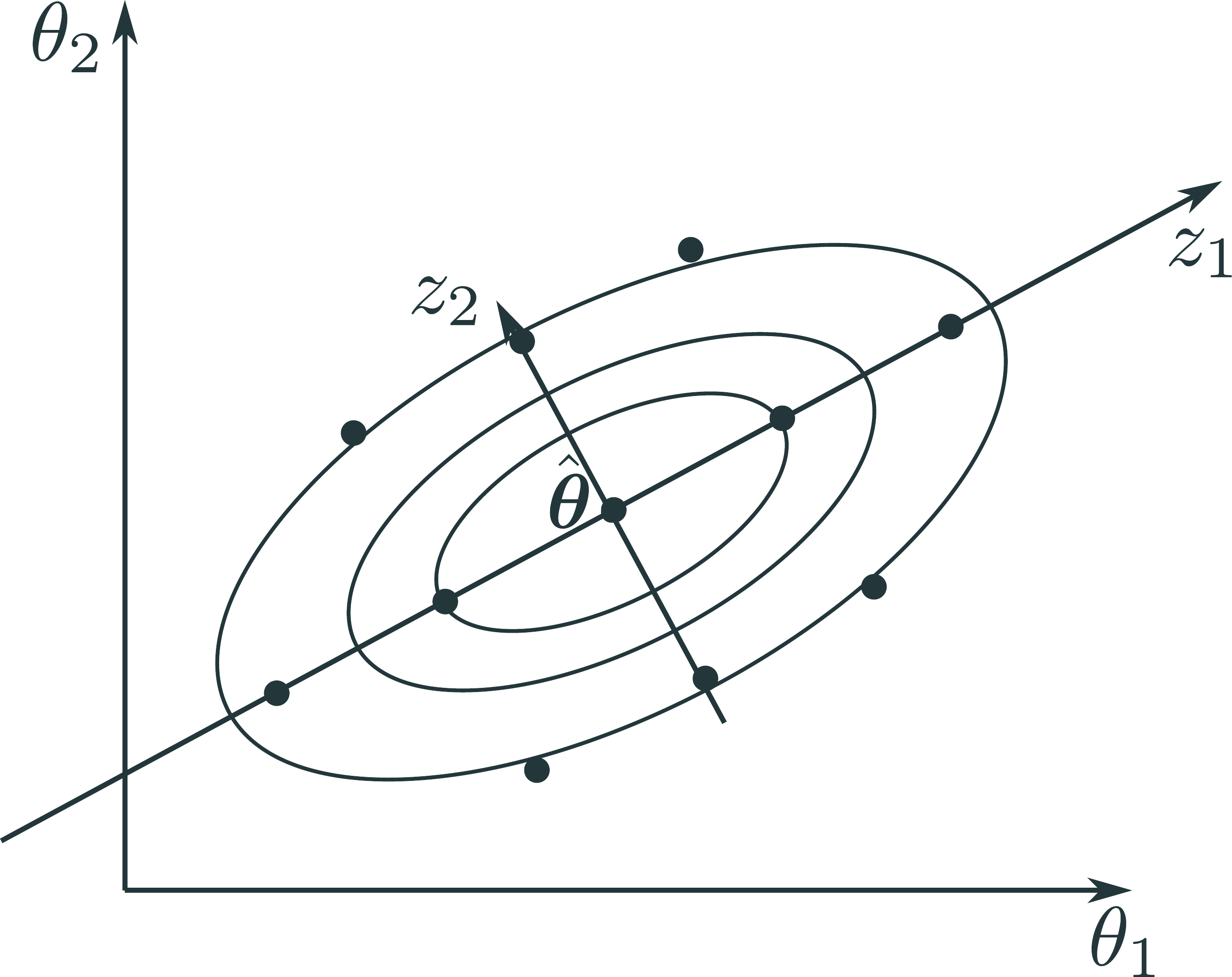}
        \label{fig:grid}
    }
    \subfigure[CCD]{
        \includegraphics[width = 0.46\columnwidth]{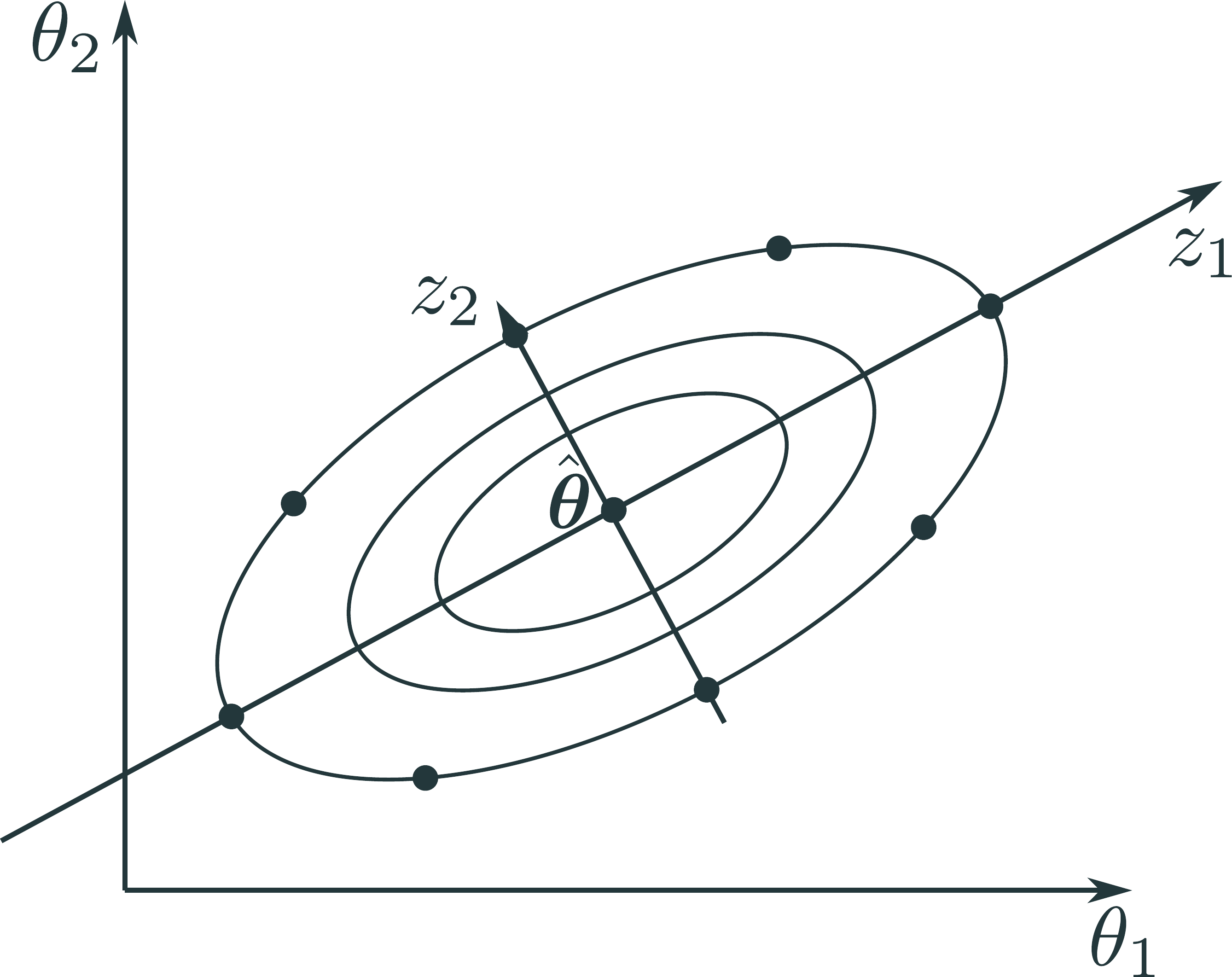}
        \label{fig:ccd}
    }
    \caption{Strategies for numerical integration of
        $\mathrm{log} ~ \tilde{p} (\boldsymbol{\theta} | \boldsymbol{y})$ when $|\boldsymbol{\theta}| = 2$.
        $\boldsymbol{z} = (z_1, z_2)^{\top}$ is the linear transformation of $\boldsymbol{\theta}$.
    If $\tilde{p} (\boldsymbol{\theta} | \boldsymbol{y})$ is Gaussian,
$\mathbb{E}[\boldsymbol{z}] = \boldsymbol{0}$ and $\mathrm{Var}[\boldsymbol{z}] = I$ are satisfied.}
    \label{fig:int_strategy}
\end{figure}

For latent variables \eqref{posterior_latent}, applying numerical integration as following:
\begin{enumerate}
    \item Exploring the mode of $\mathrm{log} ~ \tilde{p} (\boldsymbol{\theta} | \boldsymbol{y})$ using an optimization algorithm
        (e.g., quasi-Newton method) with respect to $\boldsymbol{\theta}$.
    \item Arranging integration points in accordance with grid or
        central composite design (CCD)\cite{box_experimental_1951} 
        strategy (see Fig.\ref{fig:int_strategy}). \label{enum:arrange}
    \item Calculating
        $\mathrm{log} ~ \tilde{p} (\boldsymbol{\theta}_h | \boldsymbol{y})$
        at arranged points, where $h \in \{1, \ldots, H\}$ is the index of each point.
    \item Approximating equation \eqref{posterior_latent} with weighted summation on $\theta_h$:
    \begin{IEEEeqnarray}{c}
    p(\boldsymbol{x}_i | \boldsymbol{y})
    \approx \sum^H_{h = 1} p(\boldsymbol{x}_i | \boldsymbol{y}, \boldsymbol{\theta}_h)
      p(\boldsymbol{\theta}_h | \boldsymbol{y}) \Delta_h,
    \end{IEEEeqnarray}
        where $\Delta_h$ is the weight of $\boldsymbol{\theta}_h$.
\end{enumerate}

\section{Our Model}
\subsection{ICAR Models}
To use INLA, we should choose a latent structure carefully.
This is because the computaional performance of INLA strongly depends on
the number of hyperparameters of the LGM.
\par
ICAR model is one of the commonly used representations of spatial interaction\cite{besag_conditional_1995}.
In ICAR model, $x_i$ has interaction with $x_j ~ (j \in C_i)$,
where $C_i$  indicates the set of indices of nearest neighbor nodes of $x_i$
(for simplicity, each variable is assumed to be a scalar value).
Here it assumed that $x_i$ follows
\begin{IEEEeqnarray}{c}
    p(x_i) =
    \mathcal{N} \left ( x_i \left |
    \cfrac{1}{|C_i|} \sum_{j \in C_i} x_j,
    \cfrac{\sigma^2}{|C_i|} \right . \right ),
    \label{icar}
\end{IEEEeqnarray}
where $\sigma^2$ is the hyperparameter indicates variance between nodes.
This model means some assumptions on $\boldsymbol{x}$:
\begin{itemize}
    \item Each node hardly differs from adjacent ones.
    \item The similarity between adjacent nodes is controlled by the fixed effect $\sigma^2$.
    \item As $|C_i|$ is increasing, the variance of $x_i$ decreases.
\end{itemize}
At a glance, this model seems not to be inappropriate.
However, the precision matrix of $\boldsymbol{x}$ does not satisfy
the condition of positive definiteness, then ICAR models become improper.
Therefore, we need to regularize them to guarantee its computational stabiliy.

\subsection{Proper ICAR Models}
To guarantee properness of ICAR models, two reguralization ways are frequently implemented:
scaling $\boldsymbol{x}$ by $\sum_{i} x_i = 0$ or adding extra terms to the parameters of equation \eqref{icar}.
We adopt the latter strategy, since the number of detected photons must be non-negative.
We use following proper ICAR model for the latent GMRF of adopted model,
\begin{IEEEeqnarray}{c}
    p(x_i) =
    \mathcal{N} \left ( x_i \left |
    \cfrac{1}{|C_i| + d} \sum_{j \in C_i} x_j,
    \cfrac{\sigma^2}{|C_i| + d} \right . \right ),
    \label{propericar}
\end{IEEEeqnarray}
where $d > 0$ is the extra hyperparameter for stability.
In this study, we configure interaction in a grid without periodic boundary condition
for the latent GMRF to deal with images.
For instance, we consider the probability of a non-corner pixel $p(x_l)$ of an $L \times L$ image.
Here, we can deform equation \eqref{propericar} into
\begin{IEEEeqnarray}{c}
    p(x_l) =
    \mathcal{N} \left ( x_l \left | \mu_l, \cfrac{\sigma^2}{4 + d} \right . \right ),
    \label{icarexample}
\end{IEEEeqnarray}
where
\begin{IEEEeqnarray}{c}
    \mu_l = \cfrac{1}{4 + d} ~ (x_{l-1} + x_{l+1} + x_{l-L} + x_{l+L} ).
\end{IEEEeqnarray}

\subsection{Observation Model}
In photon-limited images, the number of observated photons $y_i$ fluctuates
as following Poisson distribution with the \textit{true} value $x_i$.
Hence, we use
\begin{IEEEeqnarray}{c}
    p(y_i) = \mathrm{Poisson} (y_i | x_i)
    \label{obs}
\end{IEEEeqnarray}
as the observation model straightforwardly,
and $y_i$ is assumed to be independent from other observations with given $\boldsymbol{x}$.
Therefore, the joint distribution of $\boldsymbol{y}$ becomes
\begin{IEEEeqnarray}{c}
    p(\boldsymbol{y}) = \prod^n_{i = 1} \mathrm{Poisson} (y_i | x_i).
    \label{obs_joint}
\end{IEEEeqnarray}

\subsection{Hyperpriors}
In our model described above, we have two hyperparameters $\sigma^2$ and $d$.
As their hyperpriors, we use uninformative prior on $[0, \infty)$.

\section{Experiments}
\subsection{Preprocessing}
To handle general images with our model and control their contrast,
we firstly transform the pixel value $I_i$ in the original image into positive value $x_i$,
where the original image $\boldsymbol{I} = (I_1, \ldots, I_n)^{\top}$
and its transformation $\boldsymbol{x}$
consist of $n$ pixels.
In particular, we give linear transformation
\begin{IEEEeqnarray}{c}
    x_i = \cfrac{\lambda_{\mathrm{max}} - \lambda_{\mathrm{min}}}{I_{\mathrm{max}} - I_{\mathrm{min}}}
    (I_{i} - I_{\mathrm{min}}) + \lambda_{\mathrm{min}},
\end{IEEEeqnarray}
where $\lambda_{\mathrm{max}}$ and $\lambda_{\mathrm{min}}$ are
given parameters indicate maximum and minimum values of $\boldsymbol{x}$ respectively,
then $I_{\mathrm{max}}$ and $I_{\mathrm{min}}$ are the maximum and minimum values of $\boldsymbol{I}$.
We can control the range of intensity by varying $\lambda_{\mathrm{max}}$ and $\lambda_{\mathrm{min}}$.

\subsection{Evaluation Criteria}
In our simulations, we use peak signal-to-noise ratio (PSNR) and
structural similarity (SSIM)\cite{wang_image_2004}
to measure similarities between original and denoised images.
PSNR is defined by
\begin{IEEEeqnarray}{c}
    PSNR = 10 \mathrm{log}_{10} \cfrac{(\mathrm{max\{\boldsymbol{G}, \boldsymbol{H}\}}
    - \mathrm{min\{\boldsymbol{G}, \boldsymbol{H}\}})^2}{\mathrm{MSE}(\boldsymbol{G}, \boldsymbol{H})}\\
    MSE(\boldsymbol{G}, \boldsymbol{H}) = \cfrac{1}{n} \sum^n_i (G_{i} - H_{i})^2,
\end{IEEEeqnarray}
where $\boldsymbol{G} = (G_1, \ldots, G_n)^{\top}$ and
$\boldsymbol{H} = (H_1, \ldots, H_n)^{\top}$ denote images have $n$ pixels.
Here, PSNR just focuses on the each corresponding pixel value.
To consider the perceptual similarity between two images, we also introduce SSIM for the evaluation criterion.
SSIM is defined as follows:
\begin{IEEEeqnarray}{c}
    SSIM = \cfrac{(2\mu_G \mu_H + c_1)(2\sigma_{G,H} + c_2)}
    {(\mu_G^2 + \mu^2_H + c_1)(\sigma^2_G + \sigma^2_H + c_2)},
\end{IEEEeqnarray}
where $\mu_G$ and $\mu_H$ are the means of $\boldsymbol{G}$ and $\boldsymbol{H}$ respectively,
and $\sigma^2_G$ and $\sigma^2_H$ are the variances of $\boldsymbol{G}$ and $\boldsymbol{H}$ respectively.
Also, $\sigma_{G, H}$ indicates the covariance between $\boldsymbol{G}$ and $\boldsymbol{H}$.
In addition, $c_1, c_2$ are small positive real numbers for the stability.
\par
Here, PSNR can take from $0$ to infinity, and SSIM is in range of $-1$ to $1$.
In each criterion, the larger value is, the closer two images are.

\subsection{Simulations}
\begin{figure}[t]
    \centering
    \subfigure[Lena]{
        \centering
        \includegraphics[width = 0.29\columnwidth]{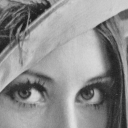}
    }
    \subfigure[Boat]{
        \centering
        \includegraphics[width = 0.29\columnwidth]{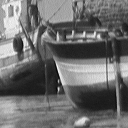}
    }
    \subfigure[Cameraman]{
        \centering
        \includegraphics[width = 0.29\columnwidth]{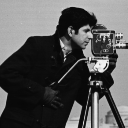}
    }
    \caption{The original images to evaluate parameter estimating methods
    for denoising Poisson corrupted images.
    Each image has been cropped into 128x128.}
    \label{fig:imgs}
\end{figure}
\begin{figure*}[t]
    \centering
    \subfigure[Lena]{
        \centering
        \begin{minipage}{0.15\textwidth}
            \centering
            original\\[0.1ex]
            \includegraphics[width = \columnwidth]{lena128.png}
        \end{minipage}
        \quad
        \begin{minipage}{0.15\textwidth}
            \centering
            corrupted\\[0.1ex]
            \includegraphics[width = \columnwidth]{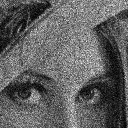}
        \end{minipage}
        \quad
        \begin{minipage}{0.15\textwidth}
            \centering
            INLA\\[0.6ex]
            \includegraphics[width = \columnwidth]{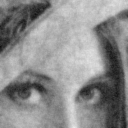}
        \end{minipage}
        \quad
        \begin{minipage}{0.15\textwidth}
            \centering
            MCMC\\[0.6ex]
            \includegraphics[width = \columnwidth]{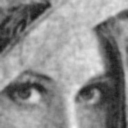}
        \end{minipage}
        \quad
        \begin{minipage}{0.15\textwidth}
            \centering
            LBP\\[0.6ex]
            \includegraphics[width = \columnwidth]{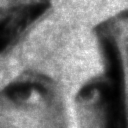}
        \end{minipage}
        \label{subfig:lena}
    }
    \\
    \subfigure[Boat]{
        \centering
        \begin{minipage}{0.15\textwidth}
            \centering
            original\\[0.1ex]
            \includegraphics[width = \columnwidth]{boat128.png}
        \end{minipage}
        \quad
        \begin{minipage}{0.15\textwidth}
            \centering
            corrupted\\[0.1ex]
            \includegraphics[width = \columnwidth]{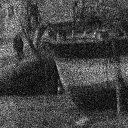}
        \end{minipage}
        \quad
        \begin{minipage}{0.15\textwidth}
            \centering
            INLA\\[0.6ex]
            \includegraphics[width = \columnwidth]{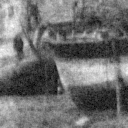}
        \end{minipage}
        \quad
        \begin{minipage}{0.15\textwidth}
            \centering
            MCMC\\[0.6ex]
            \includegraphics[width = \columnwidth]{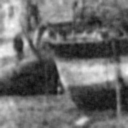}
        \end{minipage}
        \quad
        \begin{minipage}{0.15\textwidth}
            \centering
            LBP\\[0.6ex]
            \includegraphics[width = \columnwidth]{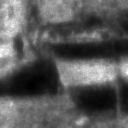}
        \end{minipage}
        \label{subfig:boat}
    }
    \\
    \subfigure[Cameraman]{
        \centering
        \begin{minipage}{0.15\textwidth}
            \centering
            original\\[0.1ex]
            \includegraphics[width = \columnwidth]{cameraman128.png}
        \end{minipage}
        \quad
        \begin{minipage}{0.15\textwidth}
            \centering
            corrupted\\[0.1ex]
            \includegraphics[width = \columnwidth]{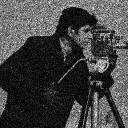}
        \end{minipage}
        \quad
        \begin{minipage}{0.15\textwidth}
            \centering
            INLA\\[0.6ex]
            \includegraphics[width = \columnwidth]{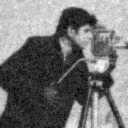}
        \end{minipage}
        \quad
        \begin{minipage}{0.15\textwidth}
            \centering
            MCMC\\[0.6ex]
            \includegraphics[width = \columnwidth]{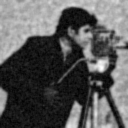}
        \end{minipage}
        \quad
        \begin{minipage}{0.15\textwidth}
            \centering
            LBP\\[0.6ex]
            \includegraphics[width = \columnwidth]{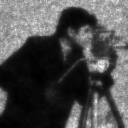}
        \end{minipage}
        \label{subfig:cameraman}
    }
    \captionsetup{width = 0.8\textwidth}
    \caption{Poisson corrupted images and denoised ones using INLA, MCMC, and LBP.
    \subref{subfig:lena}, \subref{subfig:boat}, and \subref{subfig:cameraman} shows the results of
    ``Lena'', ``Boat'', and ``Cameraman'' respectively.}
    \label{fig:denoisedimgs}
\end{figure*}
In this study, we compared denoising simulation results with INLA, MCMC
(Hamiltonian Monte Carlo with No U-Turn Sampler\cite{duane_hybrid_1987, hoffman_no_2014}),
and LBP using the described model in Section 3.
We simulated MCMC for 2000 steps and set first 1000 steps as burn-in.
As the original images, we use ``Lena'', ``Boat'', and ``Cameraman'' for the evaluation (in Fig.\ref{fig:imgs}).
Here, we fixed the contrast parameters as $\lambda_{\mathrm{min}} = 2$ and $\lambda_{\mathrm{max}} = 25$,
and ran the simulations on the computer has
AMD Ryzen Threadripper 2990WX (32 cores, 3.0 GHz) as its CPU and 128 GB memory.
To compute point estimates of $\boldsymbol{x}$,
we used expected a posteriori (EAP) estimation in INLA and MCMC.
\begin{table}[t]
    \centering
    \caption{Comparison of parameter estimating methods with PSNR, SSIM, and CPU time.}
    \begin{tabular}{c|rrrr}
        Image                      & Method & PSNR  & SSIM  & Time(s) \\\hline
        \multirow{3}{*}{Lena}      & INLA   & 22.24 & 0.966 & 226.8   \\
                                   & MCMC   & 24.97 & 0.968 & 2083.5  \\
                                   & LBP    & 21.77 & 0.928 & 18794.7 \\\hline
        \multirow{3}{*}{Boat}      & INLA   & 21.76 & 0.947 & 209.5   \\
                                   & MCMC   & 24.03 & 0.960 & 1585.3  \\
                                   & LBP    & 21.50 & 0.915 & 18891.5 \\\hline
        \multirow{3}{*}{Cameraman} & INLA   & 10.88 & 0.916 & 113.5   \\
                                   & MCMC   & 21.82 & 0.956 & 1360.7  \\
                                   & LBP    & 19.61 & 0.923 & 18809.8 \\\hline
    \end{tabular}
    \label{table:results}
\end{table}
\begin{figure*}[t]
    \centering
    \subfigure[Lena]{
        \includegraphics[width = 0.4\textwidth]{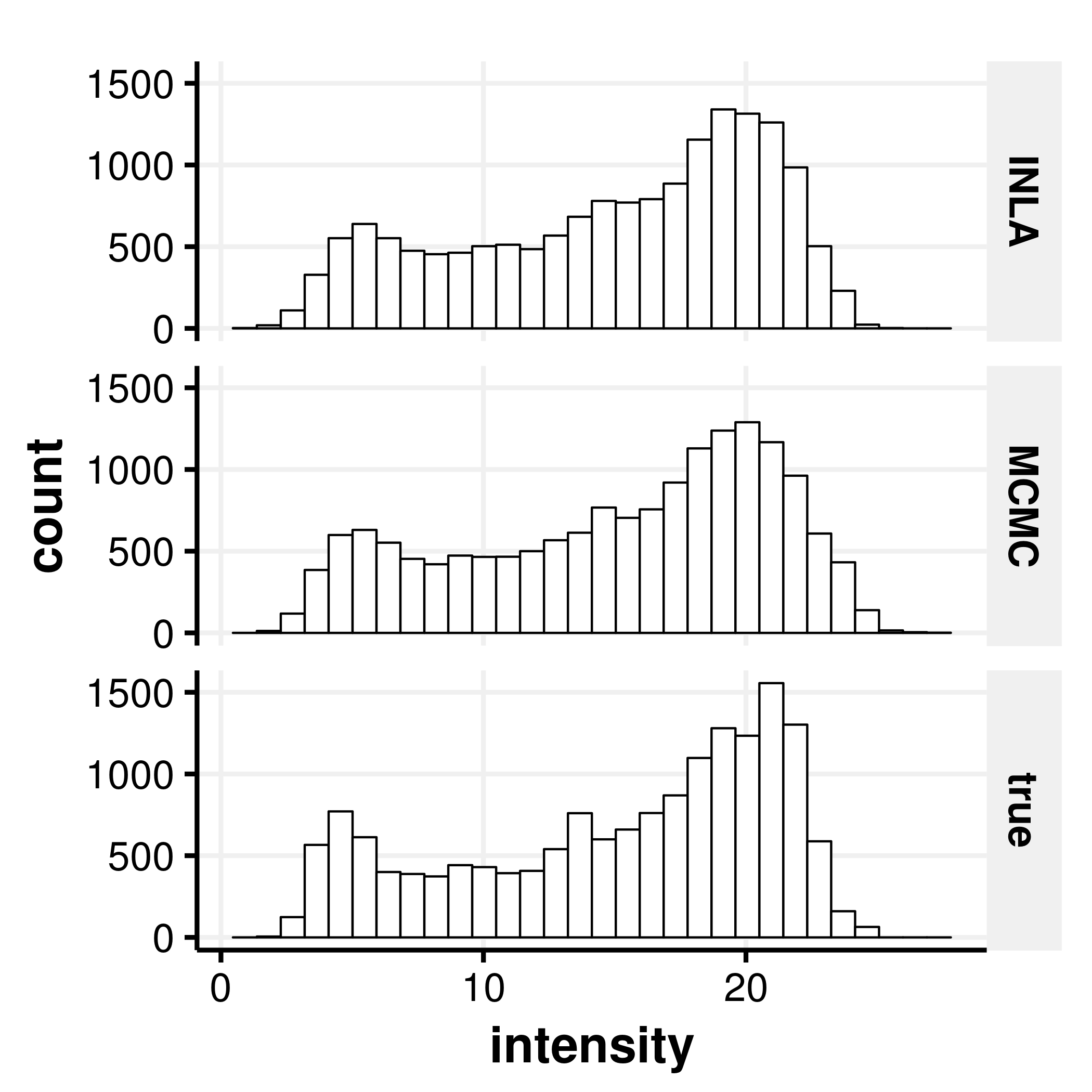}
        \label{subfig:lena_hist}
    }
    \subfigure[Cameraman]{
        \includegraphics[width = 0.4\textwidth]{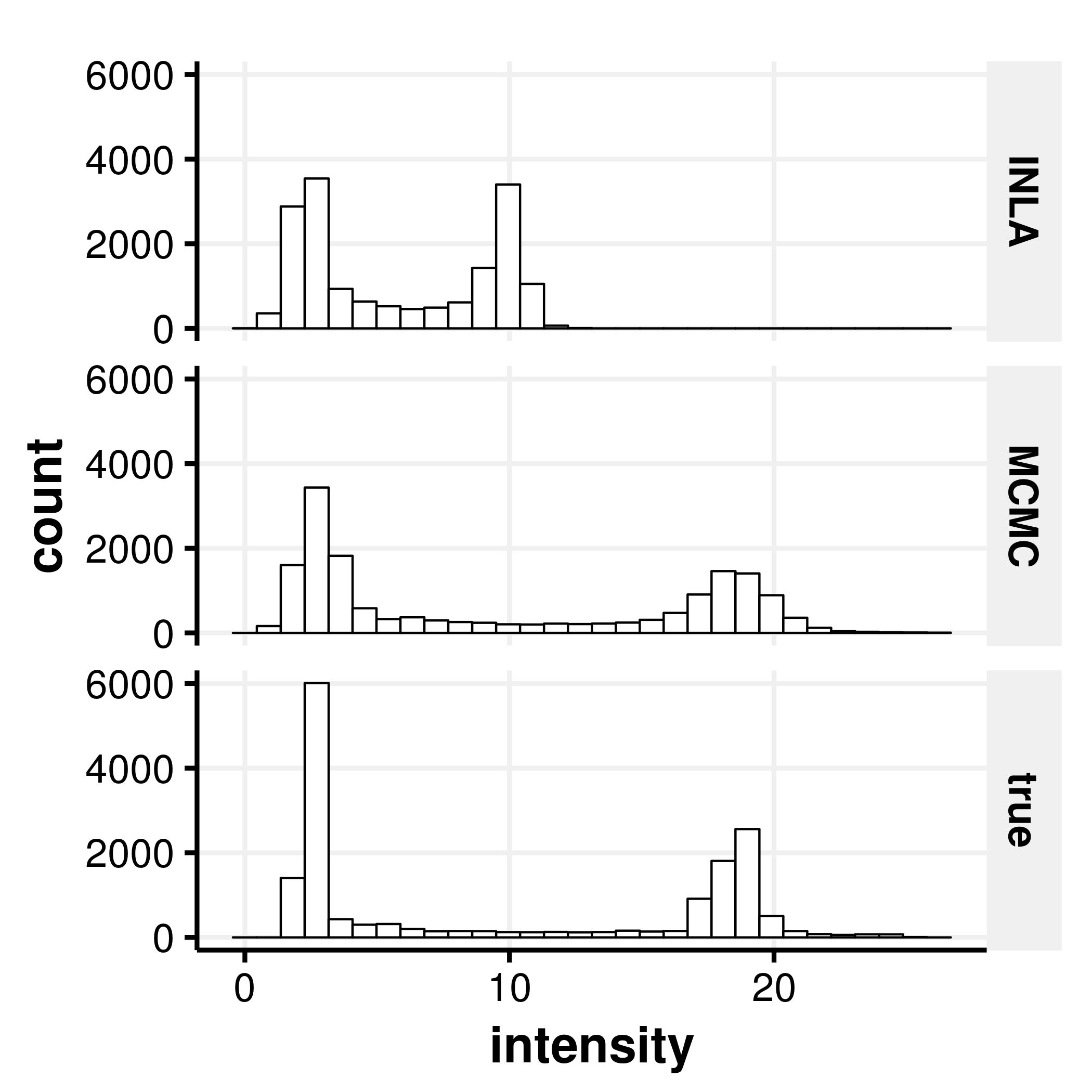}
        \label{subfig:cameraman_hist}
    }
    \captionsetup{width = 0.8\textwidth}
    \caption{The histograms of intensities of \subref{subfig:lena_hist} ``Lena'' and
    \subref{subfig:cameraman_hist} ``Cameraman''.
    Upper plot shows the estimated intensities by MCMC,
    and then middle and lower ones show INLA's and original's respectively.}
    \label{fig:hists}
\end{figure*}
Table \ref{table:results} shows the comparison of results of INLA, MCMC, and LBP.
In Table \ref{table:results}, we can find that INLA is very faster than other algorithms
with little decreasing PSNR and SSIM from MCMC, expect for ``Cameraman''.
LBP is too slow because it requires spectral decomposition of an $n \times n$ matrix in its algorithm.
Fig.\ref{fig:denoisedimgs} is the denoised images.
Qualitatively, the images restored by INLA look clearer and unhazier than by other methods on a whole.
On the other hand, we could not estimate the original ``Cameraman'' appropriately
using INLA (Fig.\ref{subfig:cameraman}).
Specifically, the result of INLA is apparently different from others in its brightness.
Then, to give additional consideration,
we present histograms of intensities in Fig.\ref{fig:hists}.
Seeing Fig.\ref{subfig:cameraman_hist}, we can find that brighter pixels have been hardly restored
in ``Cameraman''.
In ``Lena'', the shape of \textit{true} histogram is slightly flat.
In contrast, the \textit{true} histogram of ``Cameraman'' has two sharp peaks.
Therefore, it is conceivable that the error of INLA mainly caused by this illness.
In this case, INLA might fail to estimate the hyperparameters due to the illness of the original image,
then also failed to get appropriate posteriors with respect to the latent variables.
\par
\begin{figure*}[t]
    \centering
    \subfigure[Lena]{
            \centering
            \includegraphics[width = 0.45\columnwidth]{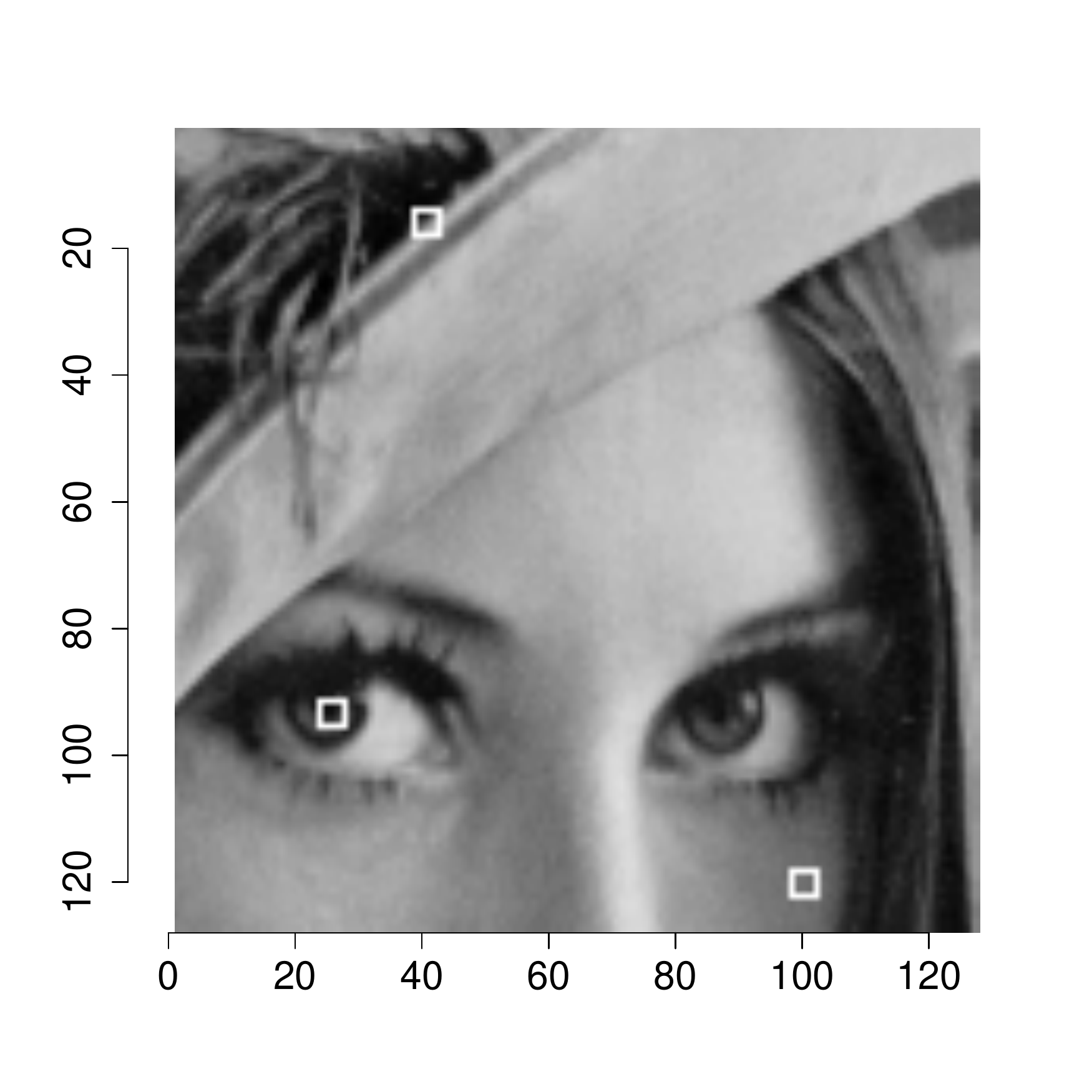}
            \includegraphics[width = 0.45\columnwidth]{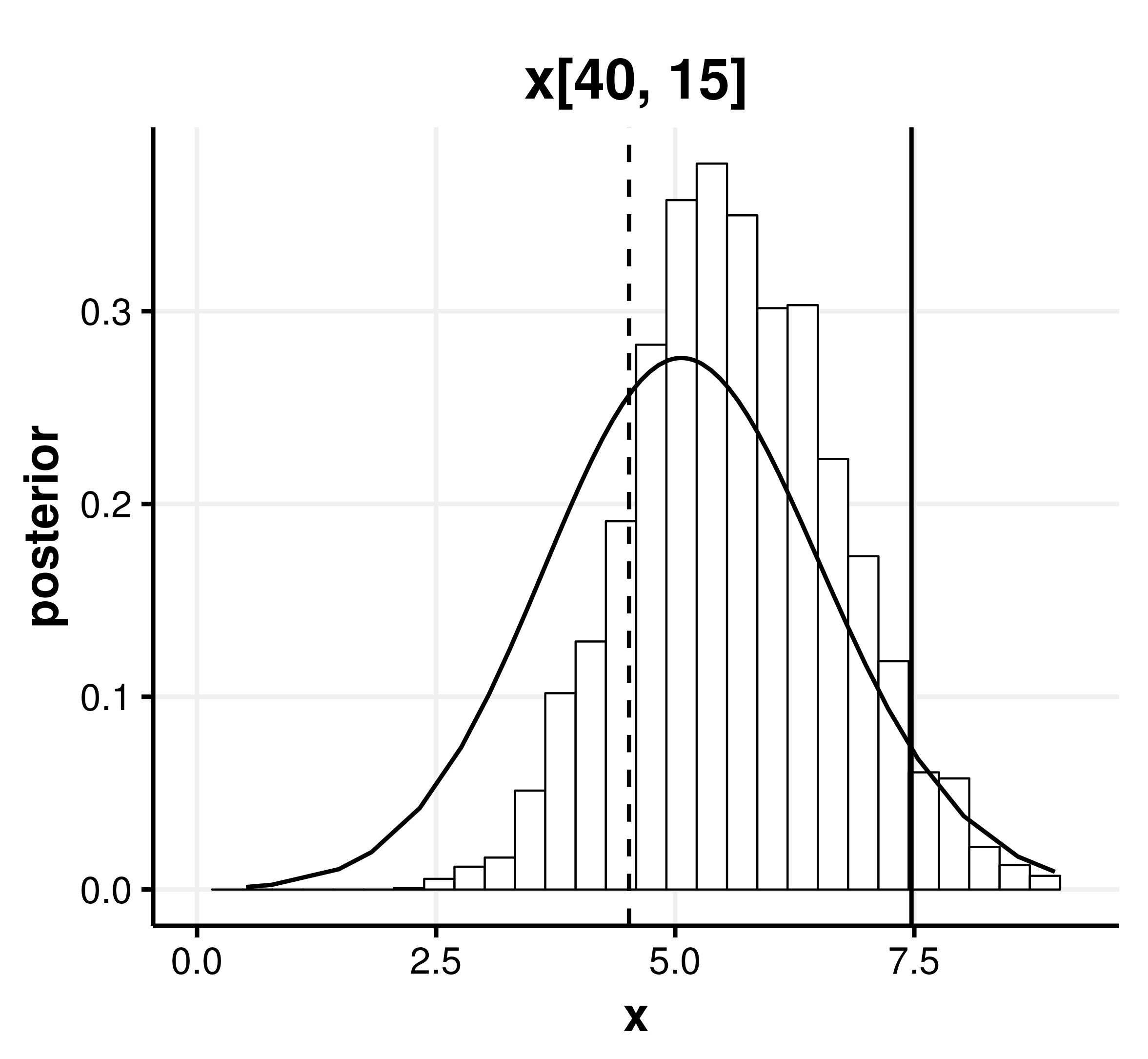}
            \includegraphics[width = 0.45\columnwidth]{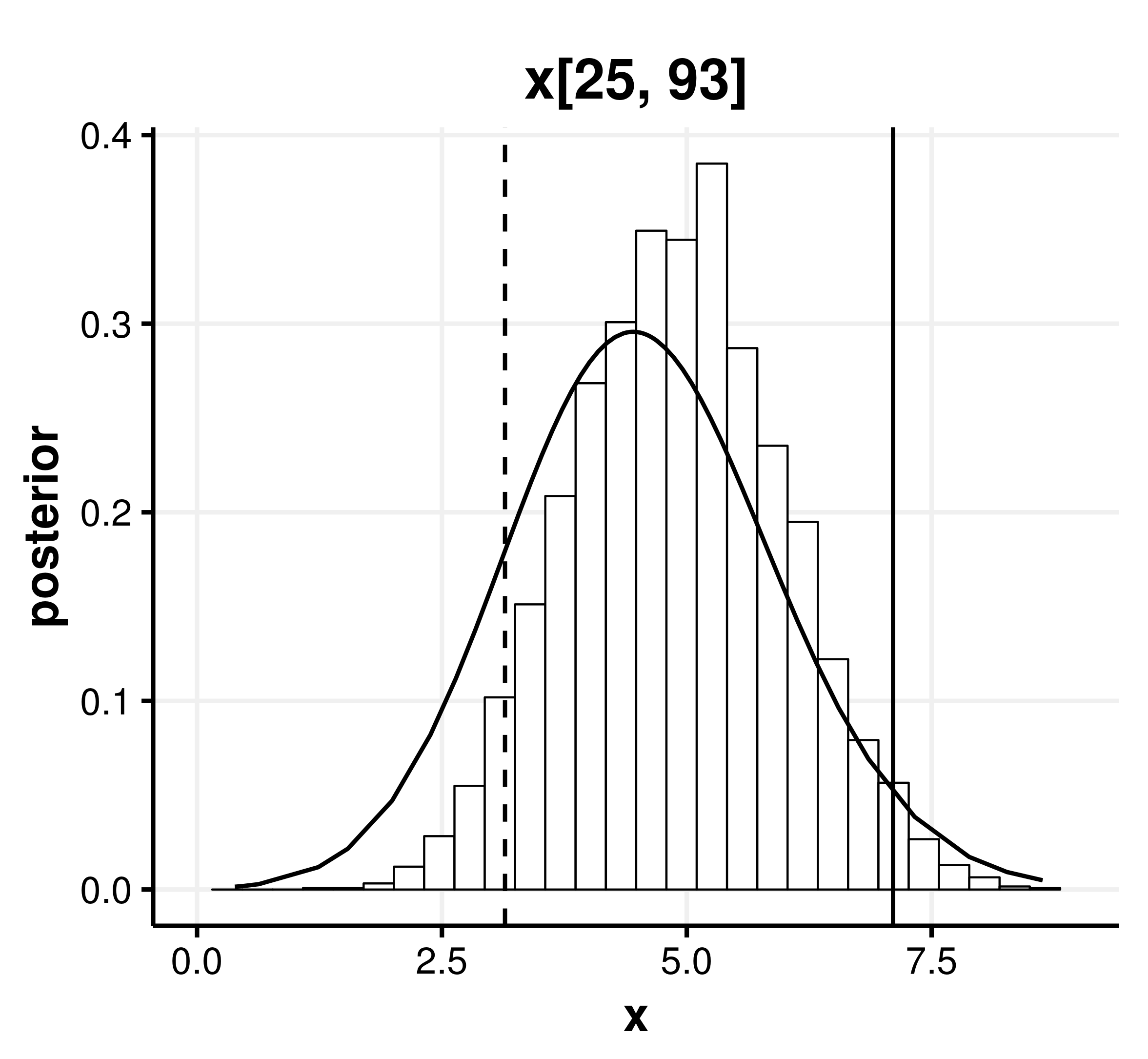}
            \includegraphics[width = 0.45\columnwidth]{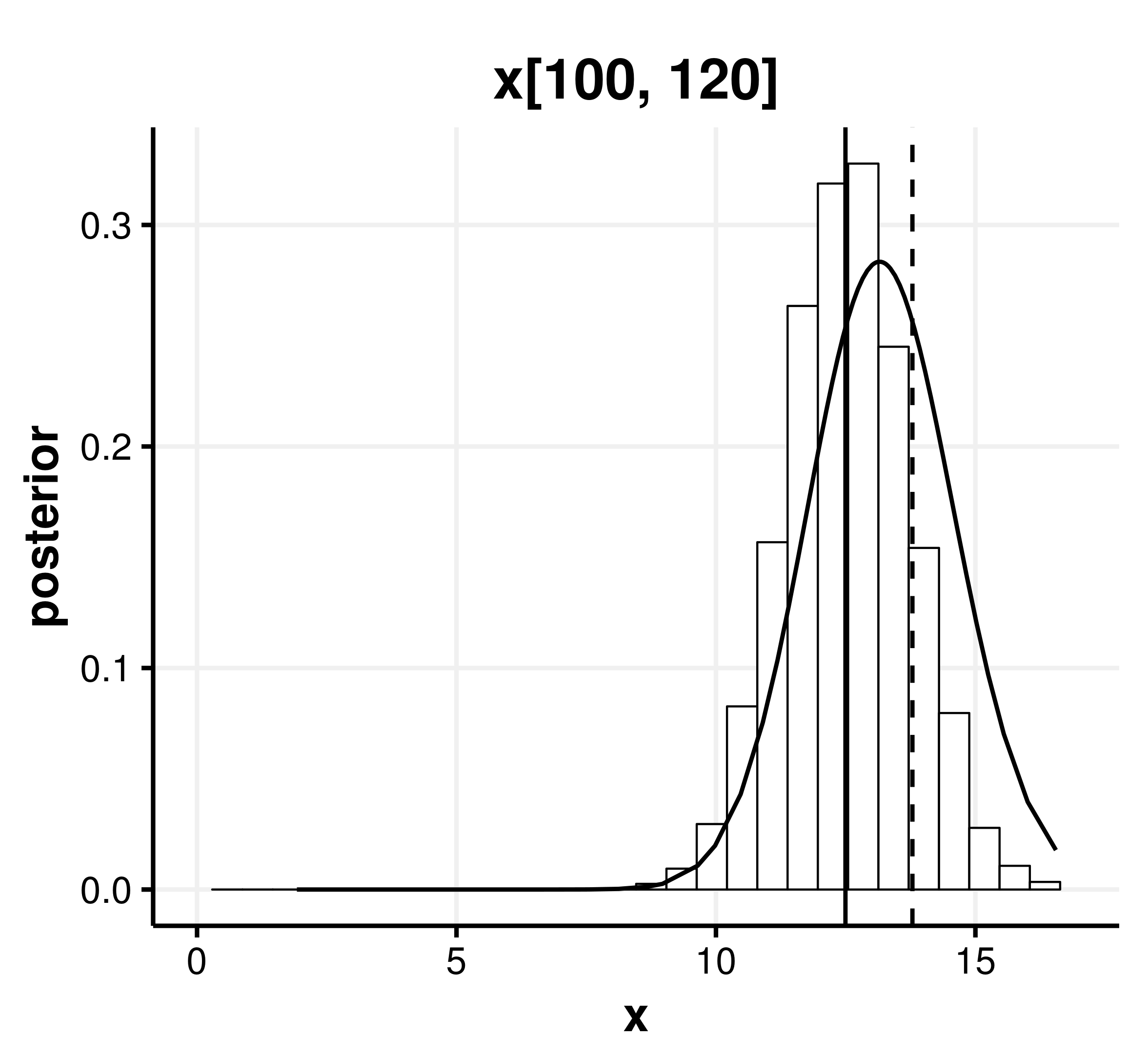}
        \label{subfig:post_lena}
    }
    \\
    \subfigure[Boat]{
            \centering
            \includegraphics[width = 0.45\columnwidth]{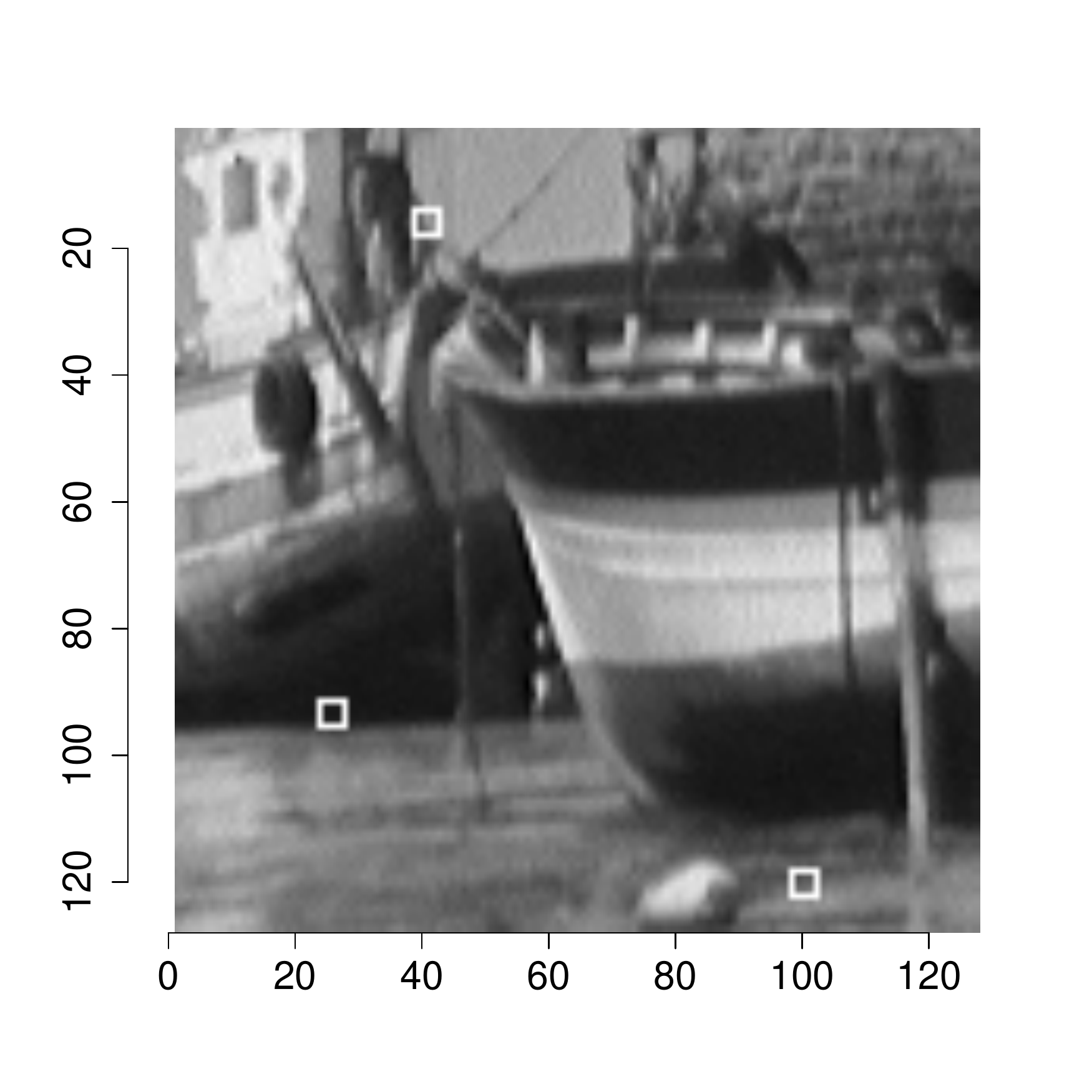}
            \includegraphics[width = 0.45\columnwidth]{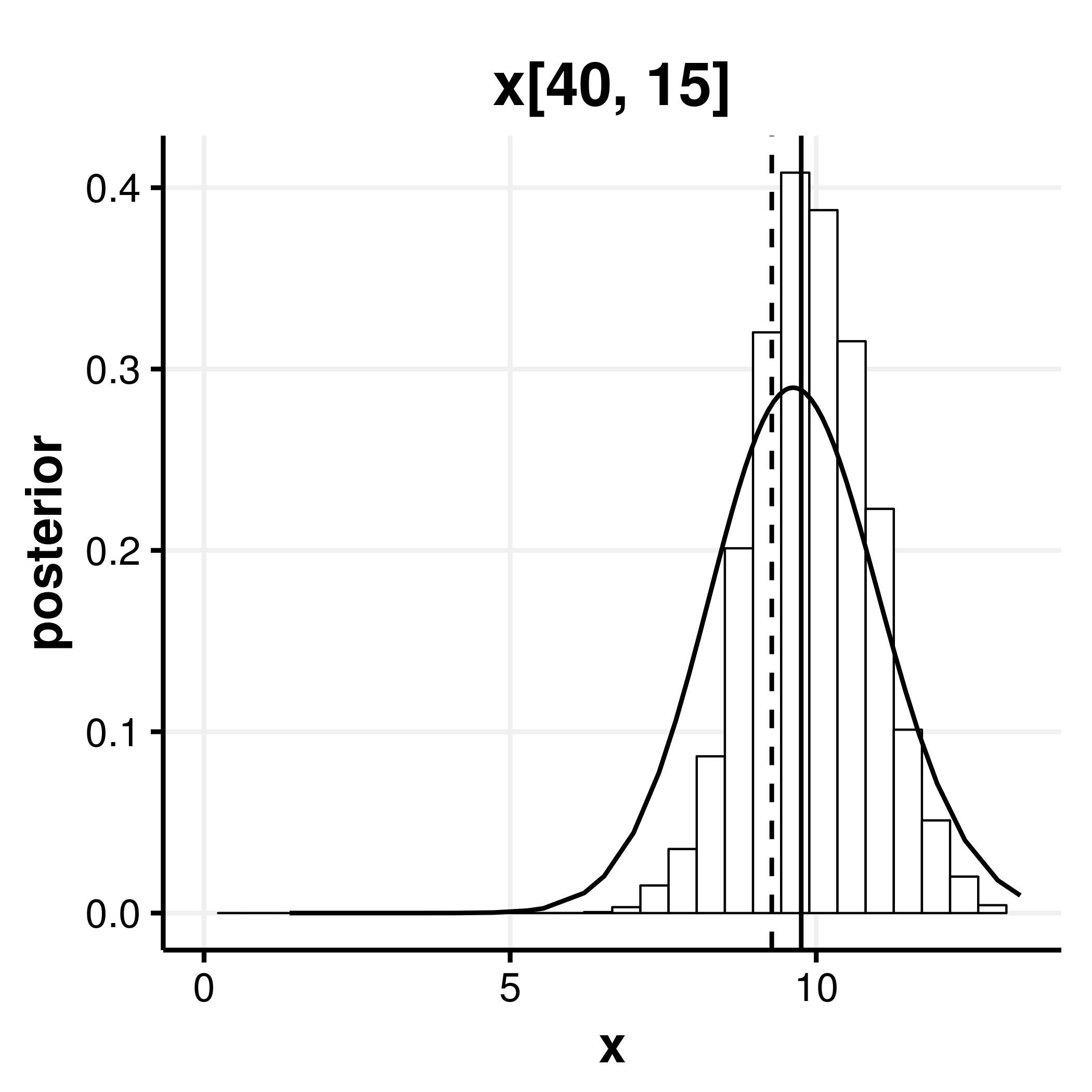}
            \includegraphics[width = 0.45\columnwidth]{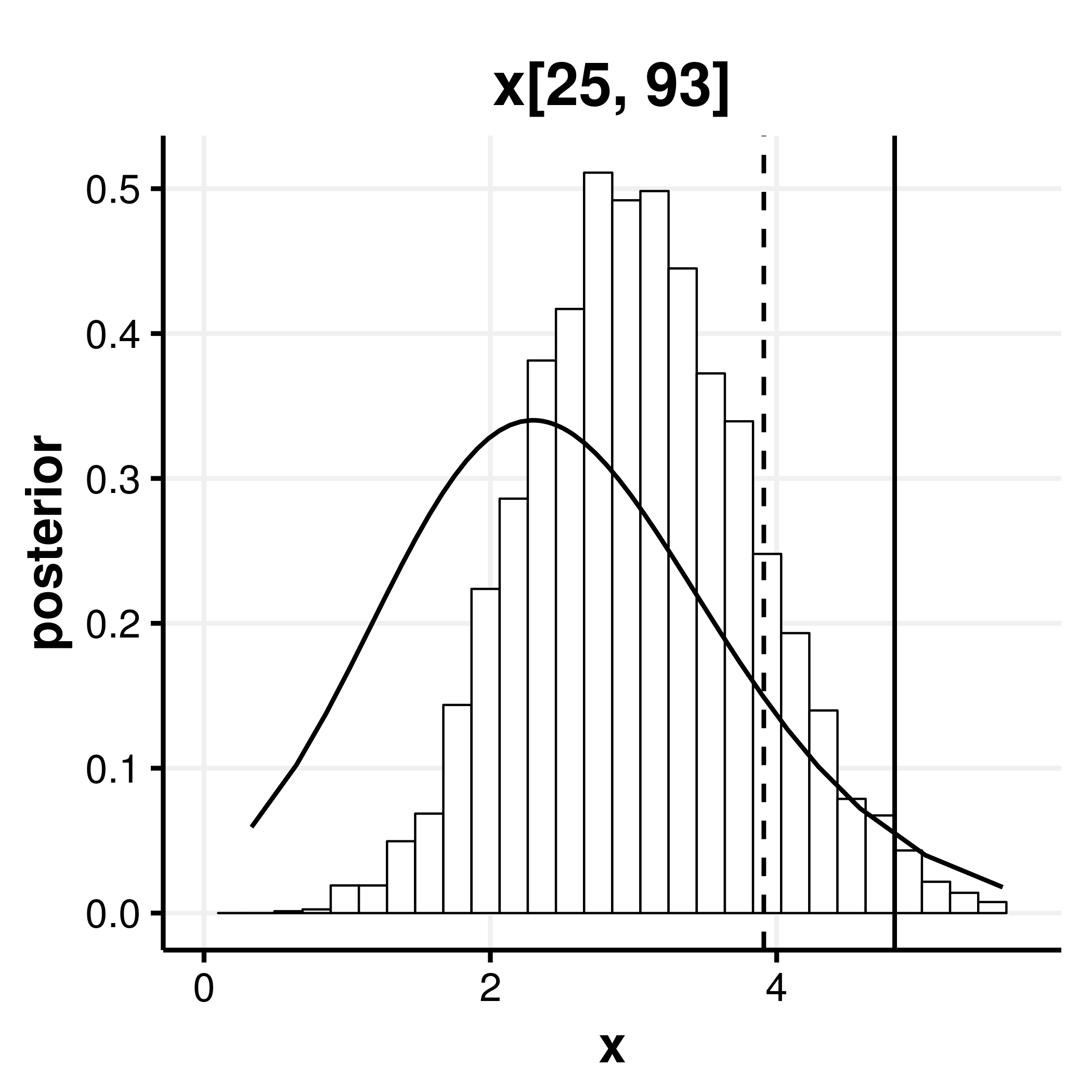}
            \includegraphics[width = 0.45\columnwidth]{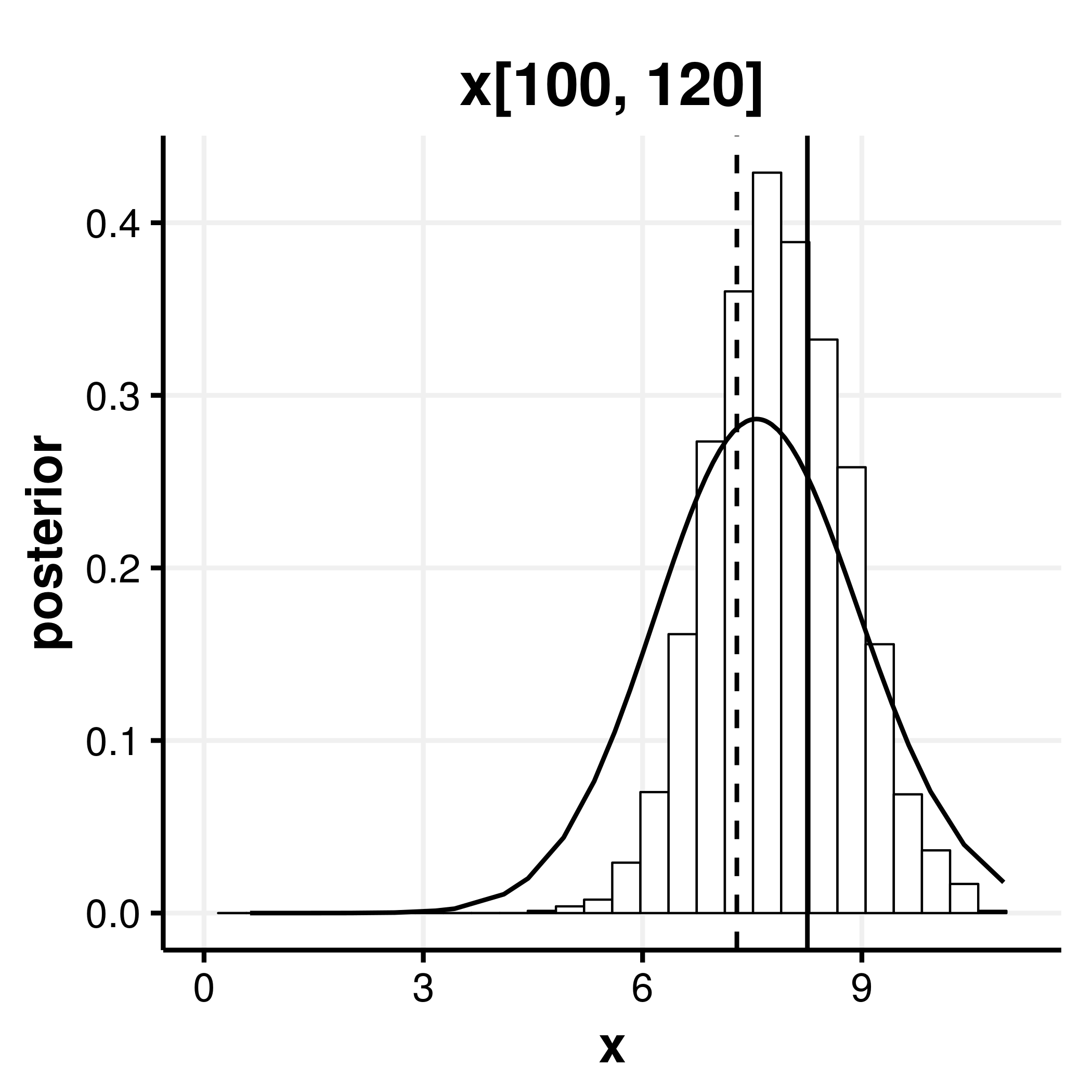}
        \label{subfig:post_boat}
    }
    \\
    \subfigure[Cameraman]{
            \centering
            \includegraphics[width = 0.45\columnwidth]{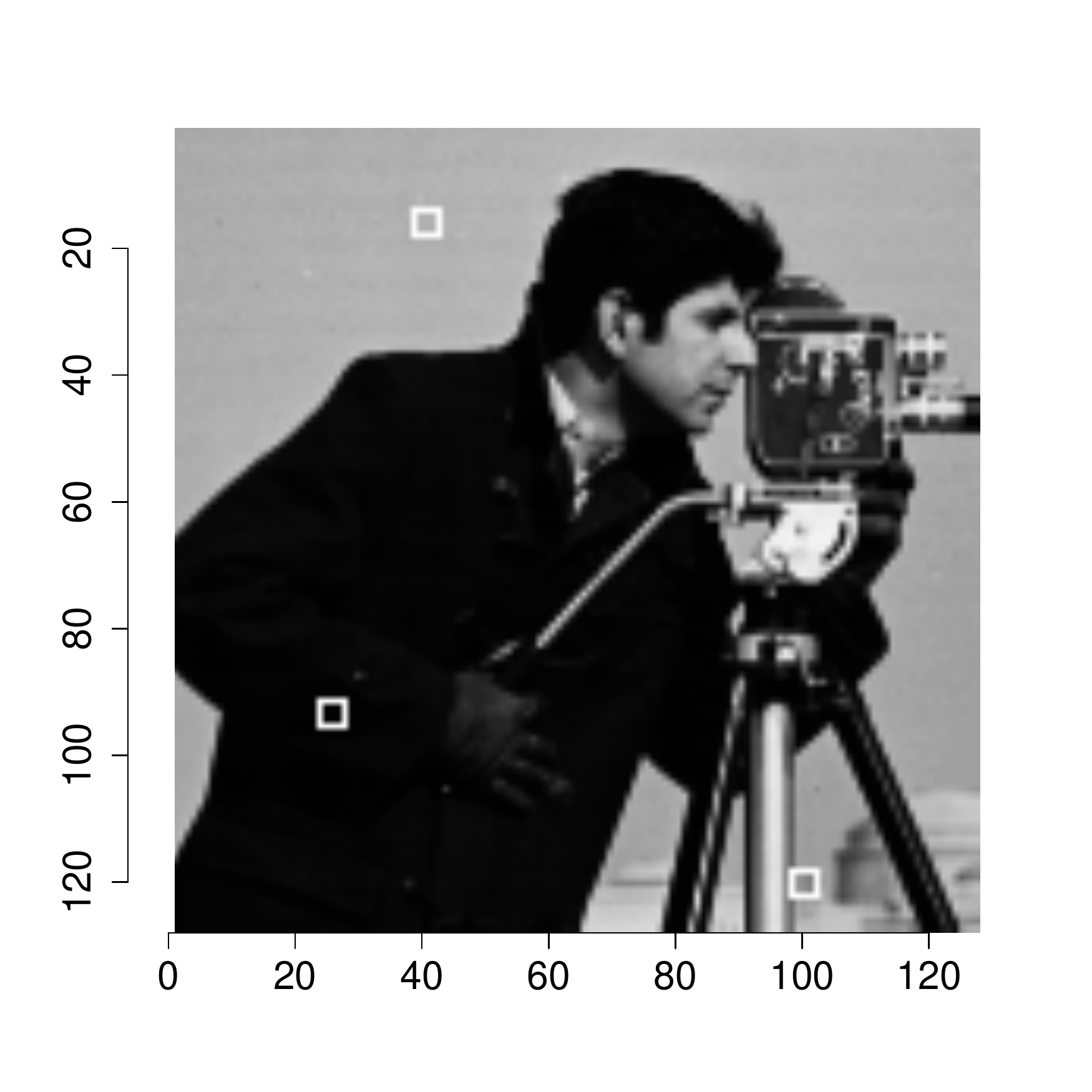}
            \includegraphics[width = 0.45\columnwidth]{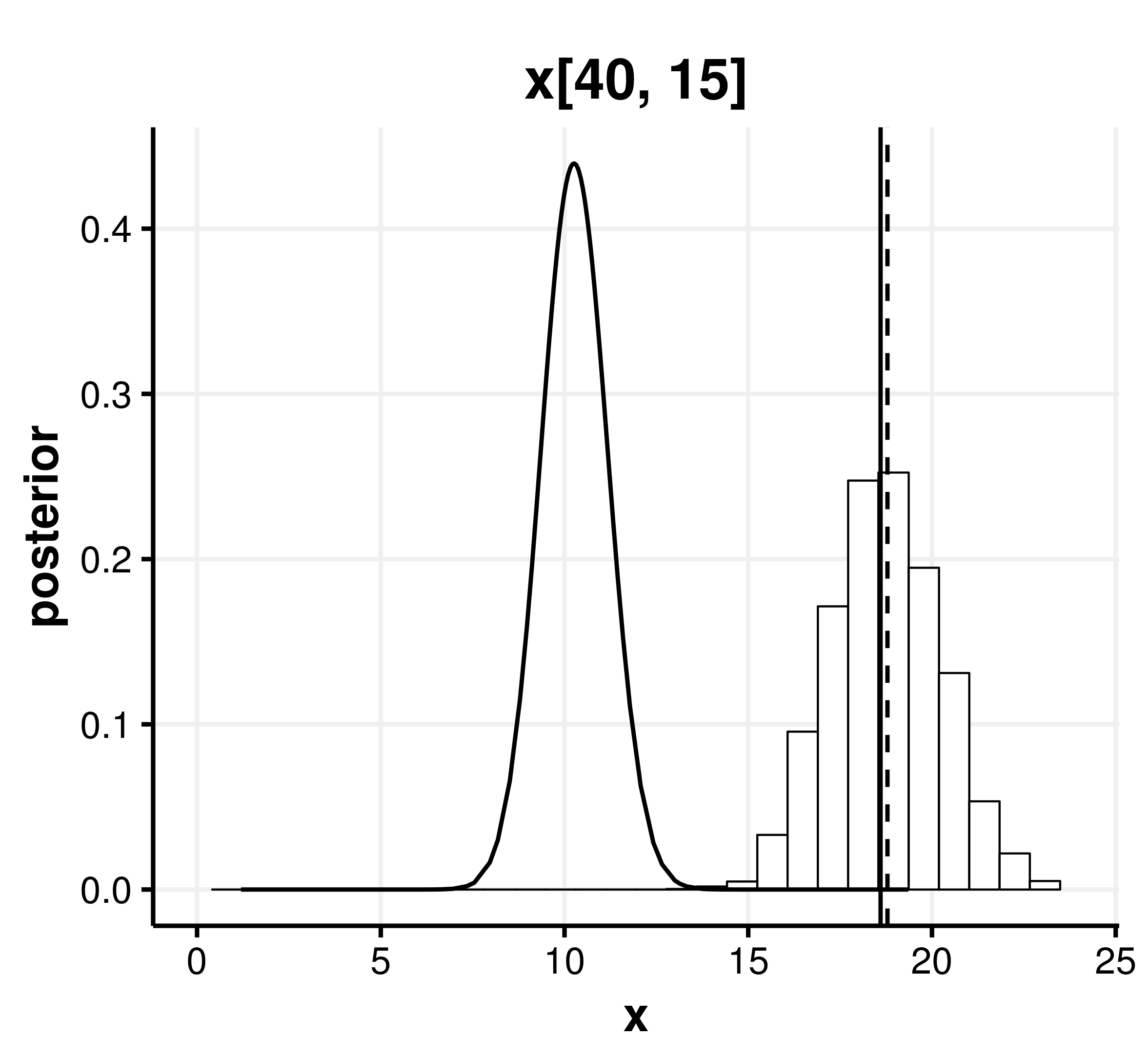}
            \includegraphics[width = 0.45\columnwidth]{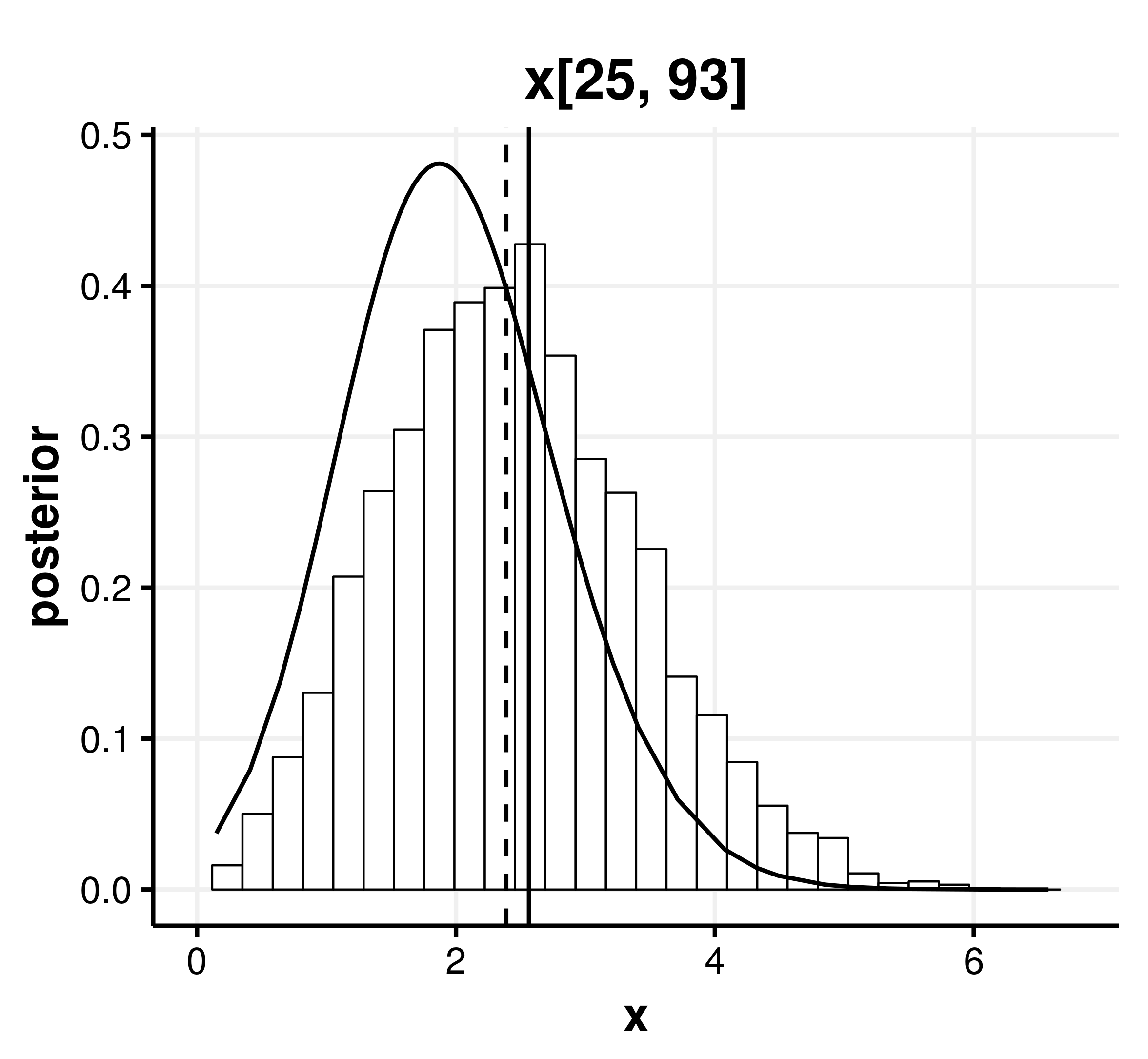}
            \includegraphics[width = 0.45\columnwidth]{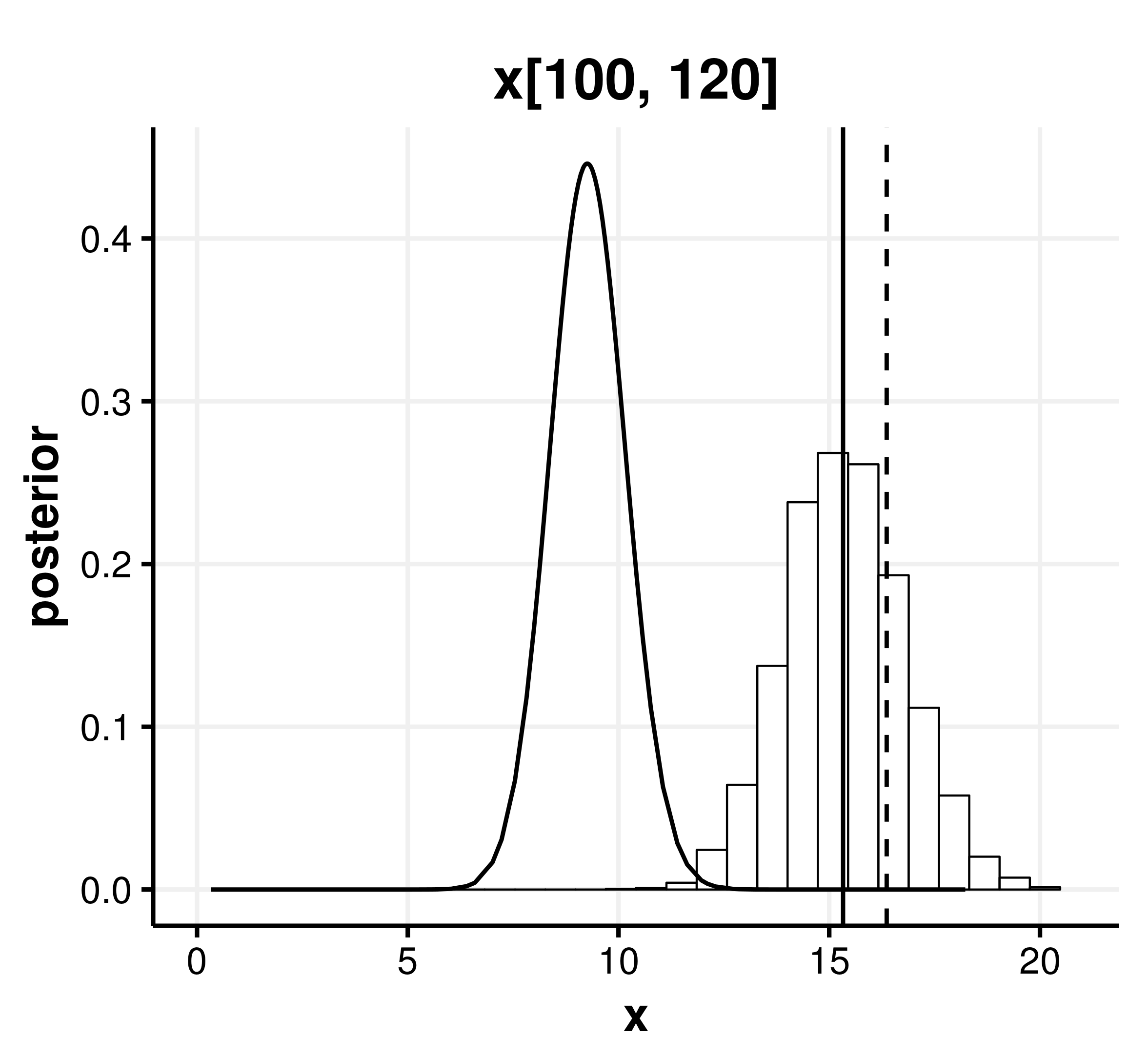}
        \label{subfig:post_cameraman}
    }
    \captionsetup{width = 0.8\textwidth}
    \caption{Comparing the posteriors (or point estimates) of some pixels of
        \subref{subfig:post_lena} ``Lena'', \subref{subfig:post_boat} `Boat'',
        and \subref{subfig:post_cameraman} ``Cameraman'' among INLA, MCMC, and LBP.
        Each picked pixel is surrounded by a square.
        Solid curves indicate approximated posteriors computed by INLA,
        and histograms show samples from MCMC.
        Then, point estimates by LBP and true values are indicated by vertical solid line and
        dashed line respectively.}
    \label{fig:posteriors}
\end{figure*}

Additionaly we indicate evaluated posteriors or its point estimates of some pixels
of each image in Fig.\ref{fig:posteriors}.
The differences of posteriors between INLA and MCMC
are unnoticeable in ``Lena'' (Fig.\ref{subfig:post_lena}).
In ``Boat'', the pixel which is near a sharp edge is slightly different the histogram by MCMC.
Moreover, we can also find that brighter pixels could not be restored in ``Cameraman''
by seeing Fig.\ref{subfig:post_cameraman}.

\section{Conclusion}
In this study, we proposed and evaluated a denoising methods for Poisson corrupted images based on INLA.
INLA requires an assumption of GMRF to latent variables.
Hence, its accuracy decreases when the assumption is inappropriate.
Whereas when the assumption is suitable for the original image,
we can reduce much time compared to other Bayesian computational algorithms,
such as LBP or MCMC with enough accuracy.
\par
As a future work, we can also consider to use other models for latent models.
For example, to apply segmentation to noisy image firstly,
and then evaluate posteriors of each segmentation independently.
In addition, we should adopt our method to real X-ray imaging data,
and then evaluate its effectivity.

%\section*{Acknowledgment}

\bibliographystyle{IEEEtran}
\bibliography{bib}

\end{document}